RESEARCH ARTICLE

# Clinical evaluation of semi-automatic open-source algorithmic software segmentation of the mandibular bone: Practical feasibility and assessment of a new course of action

Jürgen Wallner[1,2]*, Kerstin Hochegger[2,3], Xiaojun Chen[4], Irene Mischak[5], Knut Reinbacher[1], Mauro Pau[1], Tomislav Zrnc[1], Katja Schwenzer-Zimmerer[1], Wolfgang Zemann[1], Dieter Schmalstieg[3], Jan Egger[2,3,6]*

1 Department of Oral & Maxillofacial Surgery, Medical University of Graz, Auenbruggerplatz 5/1, Graz, Austria, 2 Computer Algorithms for Medicine (Cafe) Laboratory, Graz, Austria, 3 Institute for Computer Graphics and Vision, Graz University of Technology, Inffeldgasse 16c/II, Graz, Austria, 4 School of Mechanical Engineering, Shanghai Jiao Tong University, Shanghai, China, 5 Department of Dental Medicine and Oral Health, Medical University of Graz, Billrothgasse 4, Graz, Austria, 6 BioTechMed-Graz, Krenngasse 37/1, Graz, Austria

* egger@tugraz.at (JE); j.wallner@medunigraz.at (JW)

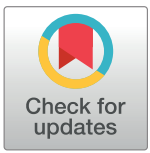

## Abstract

### Introduction

Computer assisted technologies based on algorithmic software segmentation are an increasing topic of interest in complex surgical cases. However—due to functional instability, time consuming software processes, personnel resources or licensed-based financial costs many segmentation processes are often outsourced from clinical centers to third parties and the industry. Therefore, the aim of this trial was to assess the practical feasibility of an easy available, functional stable and licensed-free segmentation approach to be used in the clinical practice.

### Material and methods

In this retrospective, randomized, controlled trail the accuracy and accordance of the open-source based segmentation algorithm GrowCut was assessed through the comparison to the manually generated ground truth of the same anatomy using 10 CT lower jaw data-sets from the clinical routine. Assessment parameters were the segmentation time, the volume, the voxel number, the Dice Score and the Hausdorff distance.

### Results

Overall semi-automatic GrowCut segmentation times were about one minute. Mean Dice Score values of over 85% and Hausdorff Distances below 33.5 voxel could be achieved between the algorithmic GrowCut-based segmentations and the manual generated ground truth schemes. Statistical differences between the assessment parameters were not

**Funding:** This work received funding from the Austrian Science Fund (FWF) KLI 678-B31: "enFaced: Virtual and Augmented Reality Training and Navigation Module for 3D-Printed Facial Defect Reconstructions" (PIs: Jürgen Wallner and Jan Egger), BioTechMed-Graz in Austria ("Hardware accelerated intelligent medical imaging"), the 6th Call of the Initial Funding Program from the Research & Technology House (F&T-Haus) at the Graz University of Technology (PI: Jan Egger) and the TU Graz Lead Project ("Mechanics, Modeling and Simulation of Aortic Dissection"). Dr. Xiaojun Chen receives support from the National Key Research and Development Program of China (2017YFB1104100, 2017YFB1302903), Foundation of Science and Technology Commission of Shanghai Municipality (15510722200, 16441908400), and Shanghai Jiao Tong University Foundation on Medical and Technological Joint Science Research (YG2016ZD01, YG2015MS26). The funders had no role in study design, data collection and analysis, decision to publish, or preparation of the manuscript.

**Competing interests:** The authors have declared that no competing interests exist.

significant (p<0.05) and correlation coefficients were close to the value one (r > 0.94) for any of the comparison made between the two groups.

## Discussion

Complete functional stable and time saving segmentations with high accuracy and high positive correlation could be performed by the presented interactive open-source based approach. In the cranio-maxillofacial complex the used method could represent an algorithmic alternative for image-based segmentation in the clinical practice for e.g. surgical treatment planning or visualization of postoperative results and offers several advantages. Due to an open-source basis the used method could be further developed by other groups or specialists. Systematic comparisons to other segmentation approaches or with a greater data amount are areas of future works.

## Introduction

In the last two decades the discipline of maxillo-facial surgery has undergone a remarkable rate of software-based technological innovation. This is especially related to the complex three dimensional anatomy of the face in combination with the need of surgical precision and an increasing number of requests for morphological three-dimensional (3D) visualized surgery [1, 2]. Therefore, the needed advanced technological and computer-based assistance is mostly based on 3D surface reconstructions or volume renderings of anatomical structures generated by segmentation algorithms [3, 4].

These segmentation approaches and segmentation algorithms are software tools and functions to be used on computer based radiological image data from computed tomography (CT) or positron emission tomography (PET/CT) scans and magnet resonance imaging (MRI) [5–9]. Due to the enlargement of medical image data in the most clinical centers–at least in the western world–, the accuracy of new image scanner generations and the low time consumption of three-dimensional (3D) image reconstruction [10], there's a rapidly growing interest in virtual segmentation automata and computer based 3D medical image analysis [4], [9], [11–14]. In the cranio-maxillofacial field these segmentation processes constitute an important step in the diagnosis and treatment planning in complex surgical cases [11, 15, 16]. The biological structures of interest in the face and skull—including soft and hard tissues—can virtually be localized, quantified and visualized to simulate 1) an interactive treatment planning, 2) complex surgical procedures and/or 3) therapeutic outcomes in three dimensions [17, 18]. Further, image based segmentation can be used to generate 3D printed models to support diagnosis and treatment pathways [15] e.g. of patients with cranio-maxillofacial deformities. If functional stable, these computer-based procedures lead to a precise preoperative representation of treatment goals, a shortened treatment or operation time and a more accurate therapeutic outcome [7, 19].

In that context an image based segmentation model created through volume rendering processes e.g. of a pathological lesions or fractured bone fragments can be created through one of the three main approaches in segmentation: manually, automatically or semi-automatically [9]. Manual image based slice-by-slice segmentation is usually tedious and time consuming [17]. Automatic segmentation algorithms based on simple thresholding and morphological operations show high sensitivity to image related artifacts which leads to a loss of segmentation accuracy through inadequate structure capturing and minor functional stability [20]. Semi-automatic, interactive segmentation approaches integrate automatic segmentation with manual guidance and are more or less a hybrid procedure of a manual and an automatic approach. Thereby, the user supports and guides the algorithm by an interactive input such as marking parts or the surroundings of the region of interest in the image to provide information of the texture and background for the software [21].

However, although robust fully and semi-automatic segmentation algorithms are described to be used in the cranio-maxillofacial complex [6], [22–25] and many interactive medical image segmentation approaches can be found in the literature as given in an overview by e.g. Zhao and Xie [26], the practical use of segmentation algorithms in clinical centers of head and neck and maxillofacial surgery is in fact hard to find. This was also observed by Egger concerning other medical fields [27].

Indeed the creation of image based segmentation models for surgical treatment planning, visualization of postoperative outcomes or for a further creation of 3D models, remains often an object of research projects without use in the practical routine or is outsourced from the clinical center to the industry as a monetary service because of e.g. personnel resources or employment reasons. This may be the reason because the algorithms presented in that field 1) do often not work stable enough, indeed fail too often or 2) are related to time consuming software processes, 3) complex user unfriendly interfaces, 4) less electronic computer powers and 5) especially to high financial costs if the software is licensed by its manufacturer. This may also be the reason that many of the major manufacturers of medical imaging equipment do not really offer sophisticated segmentation options within their workstation and software packages [27].

Having these facts in mind, little is known about the segmentation outcome and practical use of commonly available already existing and open-source based segmentation algorithms, since many authors usually present their own developed prototypes of a segmentation approach to be used for a special need, without taking respect to already existing approaches or comparing the work from other groups in experimental investigations.

Hence—due to the best of the authors' knowledge—nothing is known about the accuracy, accordance or overlap of open-source segmentation algorithms in cranio-maxillofacial surgery in direct comparison to the ground truth of the same anatomy.

Therefore, the purpose of this study was to assess the feasibility, functional stability and segmentation outcome of an open-source segmentation approach in the cranio-maxillofacial field for a further practical use in the clinical routine. Such an easily accessible digital capturing of biological structures in the cranio-maxillofacial complex through semi-automatic segmentation would further provide a better and faster morphological assessment of diagnosis and planned treatment procedures or enables the production of patient 3D models and individual implants.

In that context, the hypothesis of this study was defined as follows: The accuracy and accordance of a commonly available and licensed free semi-automatic open-source segmentation algorithm is not significantly different than the ground truth data of the same anatomy.

## Material and methods

In this retrospective, trail the accuracy and accordance of a free available segmentation algorithm was assessed through the comparison to the ground truth of the same anatomy. Therefore, the open source algorithm GrowCut (GrowCut 3.0, www.growcut.com) [28] was chosen for mandibular bone segmentation and assessment of its practical use. The foreground extraction tool GrowCut was chosen to be known as a commonly, functional stable working interactive segmentation method which is compatible with many software platforms and programs such as graphic editors. This interactive algorithm is easy to use. The user draws some strokes inside an object of interest with an object brush, and outside the object with a background brush. In simple cases, only a few strokes suffice for a complete segmentation [28]. In this trial the license-free GrowCut algorithm was carried out using semi-automatic bone countering performed on data sets of the lower jaw (**Fig 1**).

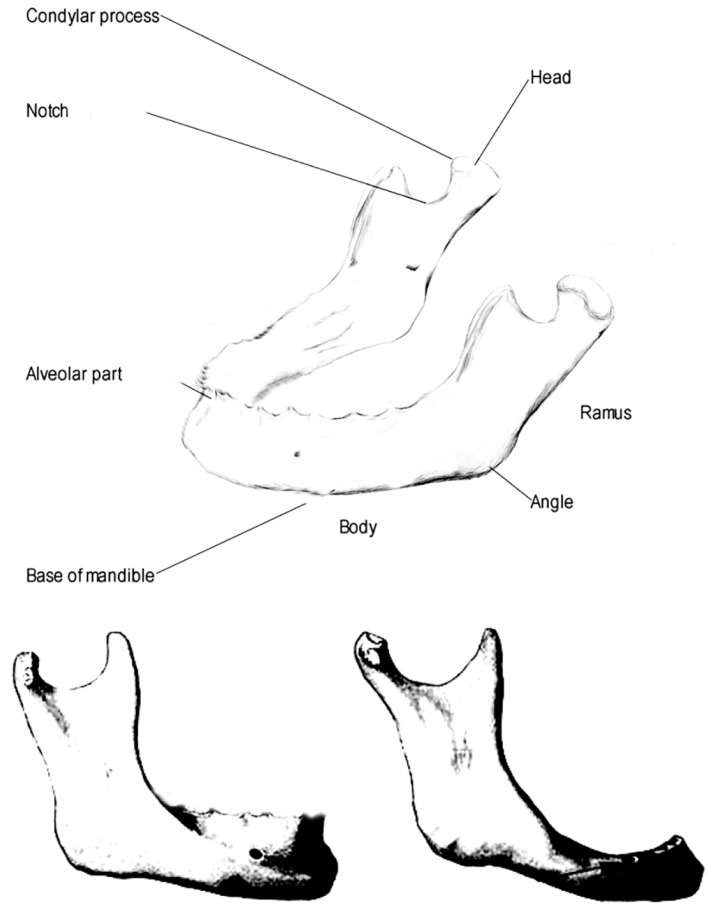

**Fig 1. The mandible: Physiologically grown anatomical structures of the mandible.** Note: There is a clear difference between the bone levels in mandibles with teeth (left) and without teeth (right).

## Data selection

For the segmentation process 45 CT-data sets were provided as DICOM files and collected during the clinical routine at the department of cranio-maxillofacial surgery at the Medical University of Graz, Austria. Only high resolution data sets (512x512) with slices not exceeding 1.0 mm with 0.25 mm pixel size and providing physiological, complete mandibular bone structures without teeth were included in the selection process. Further, no difference was made between atrophic and non-atrophic mandibular bones—both were included during the selection process. However, incomplete data sets consisting of mandibular structures altered by iatrogenic or pathological factors or fractured mandibles as well as data sets showing ostheosynthesis materials in the lower jaw were excluded in this trial. All data sets were acquired within a twelve month period (between 2013 and 2016).

According to the inclusion criteria 20 CT-data sets were selected, 25 were excluded during the selection process in the clinical routine out of diagnosis and treatment reasons. From the 20 CT-data sets, 10 data sets (n = 10, 6 male, 4 female) were further selected in a randomization process performed by a computer program (Randomizer®; https://www.randomizer.at; randomization for clinical and non-clinical trials; Graz, Austria), to form an experimental segmentation group.

The control group consisted of objective created bone structure volumes of the lower jaw according to the selected 10 data sets (ground truth). To create these ground truth volumes for a comparative assessment, a slice-by-slice manual segmentation of the randomly selected lower jaw data sets was carried out twice by two clinical experts (A, B), one specialized radiologists (A) and one specialized maxillofacial surgeon (B). More precisely, each data set was segmented manually by clinical expert A and B to create two independently ground truth schemes (Ground truth A and B) of each data set. To ensure the generation of high quality ground truth data, only physiologic data sets with clear bone contours and anatomical structures without artifacts were used in this trial, according to the mentioned inclusion criteria.

Note: All data sets were completely anonymized by the authors before their use in this study. Any patient specific information from the medical records was deleted during the anonymization process before the data selection was performed. Only de-identified data were used in this study. For this investigation we also got an approval from the internal review board (IRB) of the university (IRB: EK-29-143 ex 16/17, medical university of Graz, Austria). Since all data are within the university clinic (Medical University of Graz), the data can be retrospectively and de-identified used for research purposes and for scientific reasons. In order, the ethics committee/IRB waived the requirement for informed consent. For own research purposes the data can freely be downloaded, but we kindly asked to cite our work [29]:

https://figshare.com/articles/Mandibular_CT_Dataset_Collection/6167726

## Segmentation process

Semi-automatic segmentation using the GrowCut algorithm was carried out according to the selected data sets on the medical image computing and scientific visualization platform Slicer (Slicer 4.4.0 software (Surgical Planning Lab, Harvard Medical School, Harvard University, Boston, USA) [30], that is written in C++ and used in a variety of medical applications [31–33]. (**Figs 2A**, **2B** and **3A**) This platform is a functional stable software, easily and freely available and offers many options for medical image-based analysis such as 3D reconstructions, visualizations or preparation of 3D printable models. The automatic segmentation results performed by GrowCut could be saved by the user as a 3D mask, which was used for statistical analysis in comparison to the manual generated ground truth segmentations done by two clinical experts.

Slice-by-slice segmentation for the generation of the ground truth data was carried out on the open source scientific medical prototyping platform MeVisLab (MeVisLab 2.5.2. software (Medical imaging prototyping platform, MeVis Medical Solutions AG, Fraunhofer Institut, Bremen, Germany) [34–37] (**Fig 3B**). According to the software's function an individually created modular framework was integrated in the software platform for the ground truth generation. The MeVisLab software was used by two clinical experts (A, B) to outline the mandibular bones in the selected CT-data sets in axial directions. The selected data sets were loaded into the platforms and were successively segmented according to the used segmentation approach. After the segmentation processes were finished 10 ground truth data sets from clinical expert A, 10 ground truth data sets from clinical expert B (control group, ground truth) and 10 interactive segmentation data sets from randomly selected users were performed (segmentation group, GrowCut). The data sets were compared among themselves by defined parameters as follows: Algorithmic (GrowCut): A (Ground truth), Algorithmic (GrowCut): B (Ground truth), and A (Ground truth): B (Ground truth).

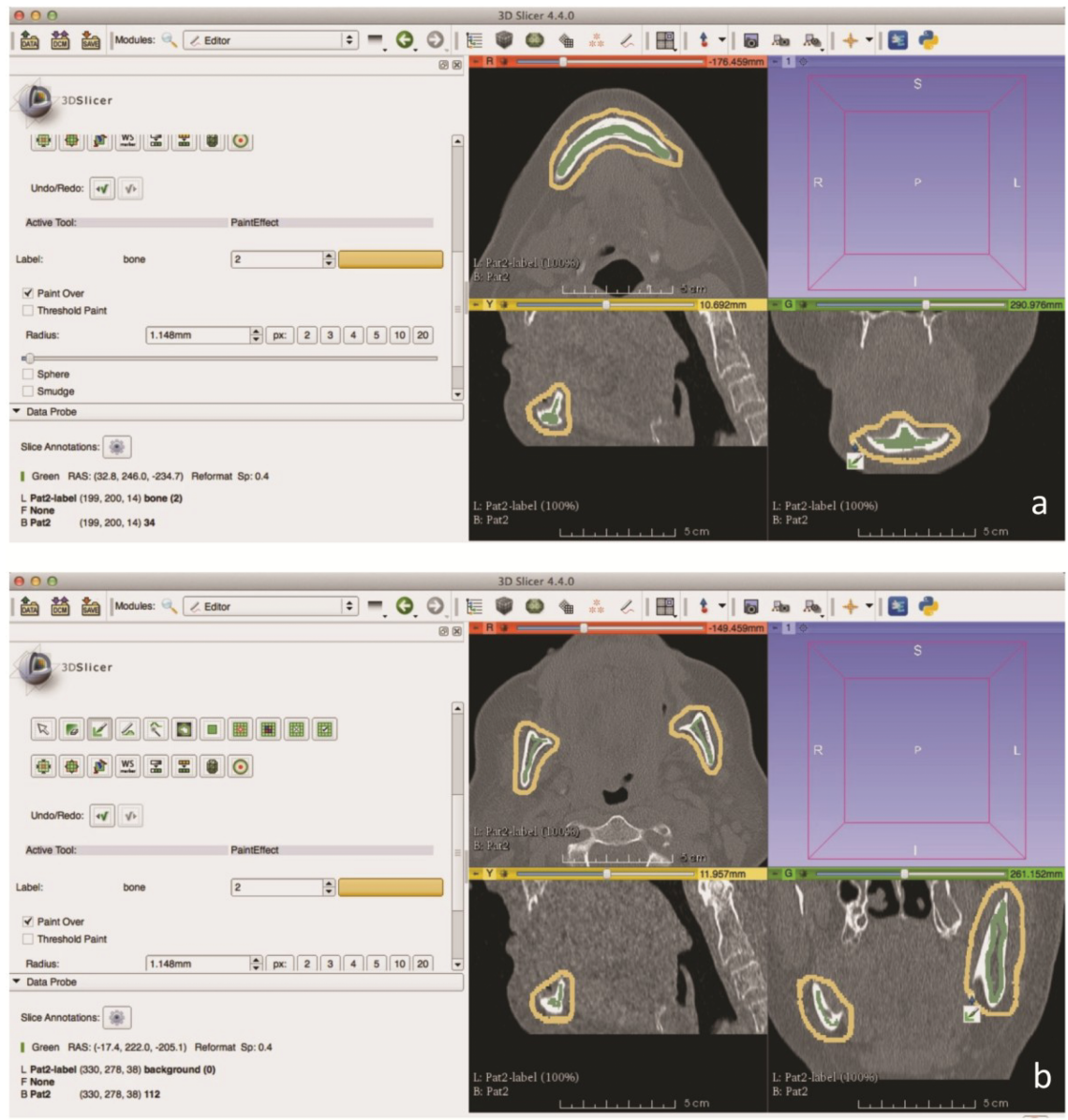

**Fig 2. Algorithmic (GrowCut) segmentation in Slicer. (a)** Fore- (green) and background (yellow) initialization of GrowCut in the lower jawbone in an axial, sagittal and coronal slice around the anterior mandible (symphysis / para-symphysis). **(b)** Slicer based algorithmic (GrowCut) segmentation: Fore- (green) and background (yellow) initialization of GrowCut in the lower jawbone in an axial, sagittal and coronal slice around parts of the mandible.

## Assessment criteria

To assess the clinical practicability of the GrowCut algorithm, randomly selected users had to initialize the GrowCut approach–after a 5 minutes introduction time–by marking parts of the mandibular bone and the background in axial, sagittal and coronal slices, respectively (Fig 2A and 2B). Each data set was segmented only once by the semi-automatic algorithm (Fig 3A).

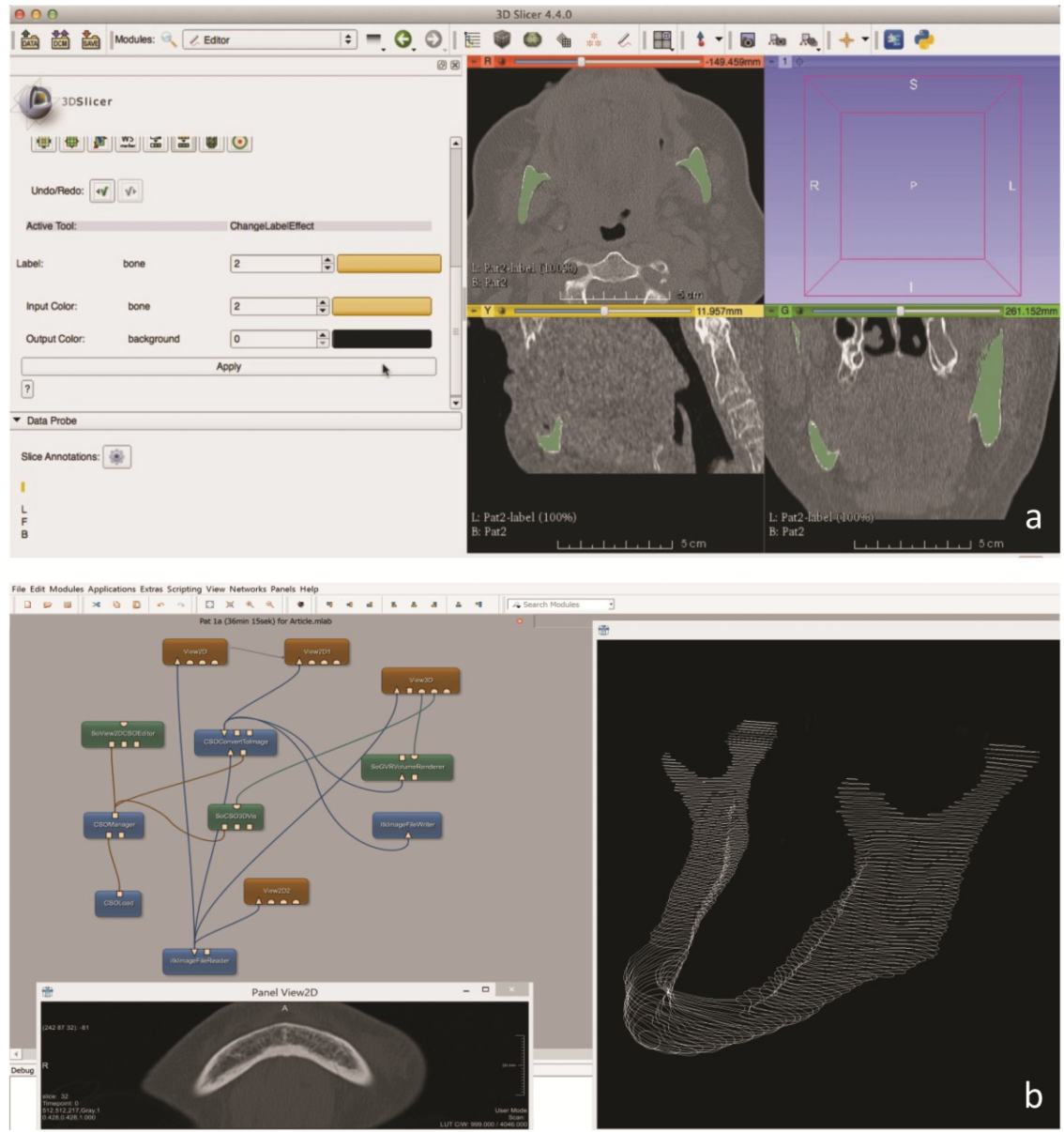

**Fig 3. (a)** Lower jaw segmentation: Final semi-automatic segmentation result (green) processed after 1 minute. **(b)** Ground truth generation: The ground truth segmentations were achieved by manual slice-by-slice segmentations by two clinical experts under MeVisLab. The screenshot shows the MeVisLab network and its modules and connections (upper left), an axial slice to draw a contour manually (lower left) and the completely segmented mandibular bone (white) in a 3D visualization (right). The single contours have been used to generate a solid 3D mask to evaluate the semi-automatic segmentations.

After segmentation, the accuracy, accordance and practicability of the semi-automatic algorithm was assessed by defined parameters. 1) The segmentation time needed was selected (starting with loading a dataset and ending with saving the single contours as one binary 3D mask) to assess the algorithm's practicability. The accuracy and accordance was assessed through the overlap between the semi-automatic open source segmentation and the ground truth of the same anatomy using 2) the DICE Score coefficient (DSC) [38, 39], 3) the Hausdorff Distance (HD) [40] 4) the segmentation volume and further 5) the number of voxel (voxel units). These parameters are known as commonly used standard sizes in the evaluation of various techniques in volume and image rendering [9, 38]. The same parameters (1–5) were assessed for the manual slice-by-slice segmentation process (ground truth A, B). All parameters were assessed for each of the 10 data sets.

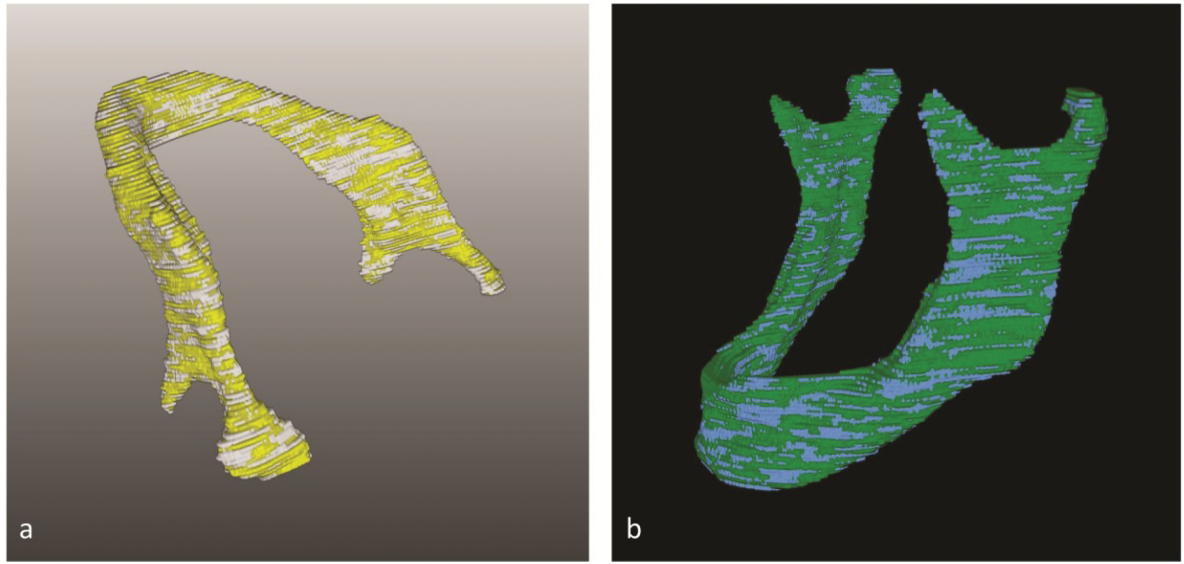

**Fig 4.** (**a**) Segmentation assessment: Overlay and accordance between the semi-automatic open-source software segmentation (yellow) and the ground truth (white) of the same structure. Calculated Dice Score, Hausdorff Distance, segmentation volume and voxel units were used as defined parameters for the assessment. *Note: Anatomical structures (condyl, incusure and others) are well visualized, by the algorithmic segmentation.* (**b**) Ground truth assessment: The ground truths' variability was assessed by comparison of the two generated ground truth schemes (blue, green) for each data set. The overlay (blue, green) of the two schemes was assesses by the Dice Score, the Hausdorff Distance, the segmentation volume and the voxel units of the anatomical structure. *Note: The ground truth was generated by manual slice-by-slice segmentation of two clinical experts (A, B).*

Each measurement (1–5) of the semi-automatic segmentation process was then directly compared to that of the two ground truth data (A, B) (**Fig 4A**). Since the ground truth data consisted of two manual segmentations performed by two specialists (A, B), these two manual segmentations were also compared among themselves (**Fig 4B**) to ensure an objective created ground truth control sample and to avoid bias causing variations in the manual segmentation.

## Statistical methods

Descriptive statistical calculations were used to summarize the measurements including minimum, maximum, mean values and standard deviations. Analytical statistical methods consisted of the calculation of paired t-tests (p) to confirm statistical validity values and the calculation of Pearson's product-moment correlation coefficient (r) [41–43], boxplots and regression analysis including regression lines through the origin. Different values were calculated between the algorithmic segmentations (GrowCut) and the ground truth data (A, B), as also between the ground truth data themselves. P-values under 0.05 (p<0.05) were assumed to be significant. All statistical calculations were performed using the R software (R-project Ⓡ; R Foundation for Statistical Computing, Vienna, Austria, ver. 3.1.2).

Figs 5 and 6 give a stepwise overview about the trail's workflow procedures including parts of the assessment criteria and the investigated algorithmic segmentation process.

## Results

The goal of this study was to evaluate the feasibility and assessment of an algorithm-supported jawbone open source software segmentation for the clinical practice. In doing so, two metrics were used for an directly in-depth evaluation of the GrowCut algorithm: the agreement between two segmentations (manual A: manual B and manual A, B: algorithmic), expressed as Dice Score and Hausdorff Distance and further volume values, voxel units and the segmentation time needed.

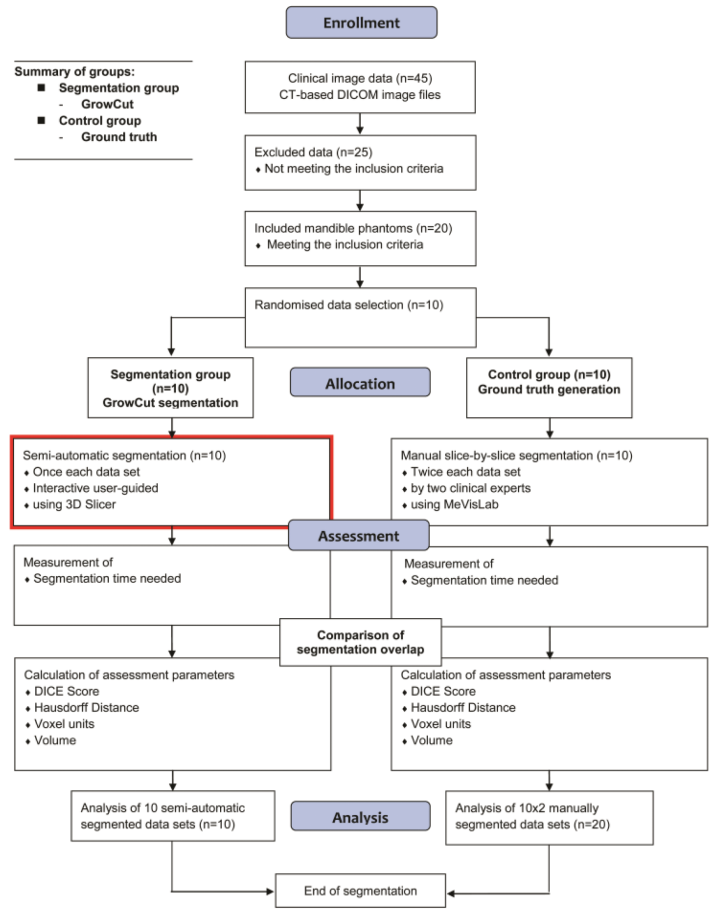

**Fig 5. Workflow: Details of the performed procedures are shown stepwise including data selection and assessment criteria.** The red box marks the segmentation process performed with the GrowCut algorithm.

A detailed analysis of the assessed parameters is provided in the **Tables 1** to **6** for every case (1–10).

**Table 1** compares the ground truth segmentation performed by two clinical experts directly and **Table 2** presents a descriptive statistical analysis of these segmentations including minimum, maximum, mean values and standard deviations. **Tables 3** and **4** compares the manual ground truth segmentation performed by clinical experts A and B directly to the results of the semi-automatic GrowCut segmentation. **Table 5** directly compares the number of voxel of the Ground truth scheme A, B and the semi-automatic segmentation (GrowCut), including minimum, maximum, mean values and standard deviations. **Table 6** summarizes the overall segmentation results of manual (ground truth) and algorithmic GrowCut segmentation minimum, maximum, mean values and standard deviations.

Overall the semi-automatic segmentation performed by the GrowCut algorithm was easy to initialize and provides an accurate segmentation in a short time in every case. All data sets

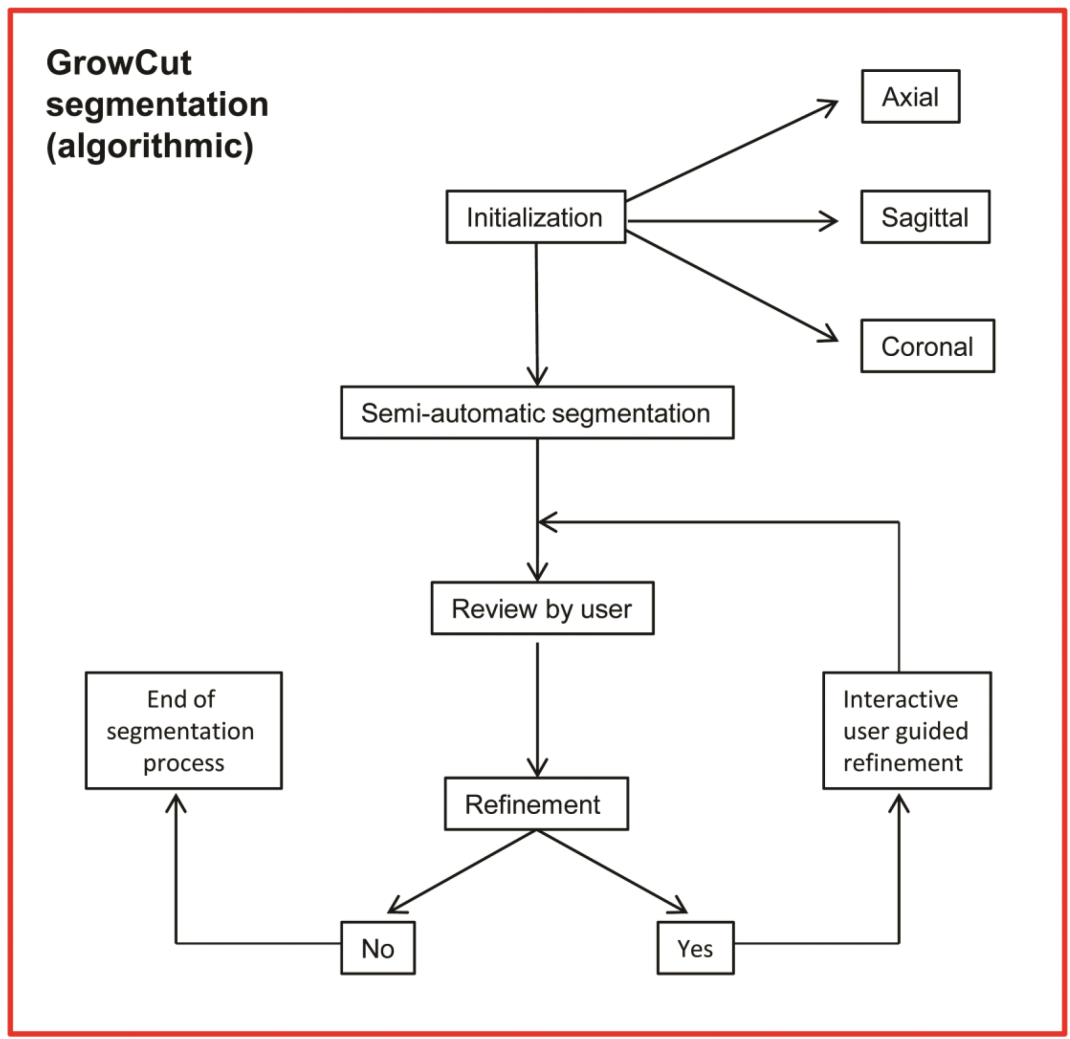

**Fig 6. Workflow: The phases of the user guided GrowCut segmentation process is shown separately in more detail.** Note: The used segmentation procedures were performed equally for each data set.

https://doi.org/10.1371/journal.pone.0196378.g006

**Table 1. Direct comparison between manual slice-by-slice ground segmentations performed by two clinical experts (A, B) for all ten lower jawbones.** DSC: Dice Score Coefficient (%), HD: Hausdorff Distance (voxel).

| Lower Jawbone No. | Volumes of the lower Jawbones (mm³) | | HD(voxel) | DSCs (%) | Times (min.)Ground truth A \| Ground truth B | |
|---|---|---|---|---|---|---|
| | Ground truth A | Ground truth B | | | | |
| 1 | 30507.8 | 29413.4 | 3.16 | 94.33 | 36 | 40 |
| 2 | 17333 | 17730.4 | 5.2 | 91.72 | 46 | 40 |
| 3 | 19356.9 | 20067.2 | 3.16 | 92.65 | 38 | 39 |
| 4 | 46506.9 | 47508.8 | 6.32 | 94.66 | 38 | 38 |
| 5 | 39813.6 | 39733 | 3.32 | 93.68 | 37 | 35 |
| 6 | 30861.2 | 31283.1 | 4.12 | 94.48 | 43 | 40 |
| 7 | 45792.7 | 45492.8 | 4.69 | 94.11 | 38 | 42 |
| 8 | 31525.1 | 32288.9 | 2.24 | 94.23 | 36 | 37 |
| 9 | 18150.5 | 18686.3 | 4.24 | 92.53 | 38 | 38 |
| 10 | 32951.8 | 31296.5 | 3.46 | 93.73 | 36 | 35 |

https://doi.org/10.1371/journal.pone.0196378.t001

**Table 2. Summary of manual vs. manual (Ground truth A, B) segmentation results, presenting minimum (min), maximum (max), mean ($\mu$) values and standard deviation ($\sigma$) for ten lower jawbones.** DSC: Dice Score Coefficient (%), HD: Hausdorff Distance (voxel).

| | Volumes of the lower Jawbones (cm$^3$) | | Ground truth A vs. Ground truth B | | Times (min.)Ground truth A \| Ground truth B | |
|---|---|---|---|---|---|---|
| | Ground truth A | Ground truth B | HD(voxel) | DSC (%) | | |
| Min | 17.33 | 17.73 | 2.24 | 91.72 | 36 | 35 |
| Max | 46.51 | 47.51 | 6.32 | 94.66 | 46 | 42 |
| $\mu\pm\sigma$ | 31.28±10.69 | 31.35±10.59 | 3.99±1.18 | 93.61±0.98 | 38.6±3.31 | 38.4±2.27 |

https://doi.org/10.1371/journal.pone.0196378.t002

were successfully segmented by the user guided interactive algorithm without interruptions. The users introduced in the investigated segmentation task were able to perform it in around one minute and successfully saved the segmented volume as a 3D binary mask. These segmentation times were far below compared to the manual slice-by-slice segmentation which were 38.6±3.31 minutes (Ground truth A) and 38.4±2.27 minutes (Ground truth B) minutes on average (**Tables 1–4**). Mean volume values of manual slice-by-slice segmentations were 31.28 ±10.69 cm$^3$ (Ground truth A) and 31.35±10.59 cm$^3$. In comparison semi-automatic GrowCut segmentations were measured to be 32.18±13.02 cm$^3$ on average (**Tables 1–4** and **6**). Similar relations were observed when comparing the number of voxel of the segmented data sets. On average these values were 119147±46957.5 for the Ground truth A, 119200.7±45568.9 for the Ground truth B and 123613.5±58013.4 for the interactive GrowCut segmentation (**Table 5**).

The overlap agreement between two manual segmentations (A, B) yielded to a Dice Score of 93.61±0.98% and the agreement between a manual and an automatic segmentation yielded to a Dice Score of 85.46±3.38% (Ground truth A) and 85.75±3.39% (Ground truth B). The calculated Hausdorff Distances were 3.99±1.18 voxel units between two manual segmentations (A, B). Further, the Hausdorff Distances were 33.51±13.98 voxel units between the manual ground truth A and the semi-automatic segmentation and 33.43±13.86 voxel units between the manual slice-by-slice segmentation B and the GrowCut algorithm (**Table 6**).

Calculated difference values between the two groups were not significant for the assessment parameters volume and voxel, neither between the manual ground truth schemes nor between the manual and the semi-automatic segmentations (p<0.05) (**Tables 7** and **8**). Further, the product-moment correlation coefficient (Pearson) was not below 0.94 (r>0.94) in every case when the two groups were compared with each other (**Tables 7** and **8**).

**Table 3. Direct comparison of manual slice-by-slice (Ground truth A) and Slicer-based semi-automatic GrowCut segmentation (algorithmic) results for ten lower jawbones via the Hausdorff Distance and the Dice Similarity Coefficient (DSC).** The last column presents the time in minutes (min.) for the manual segmentations; in contrast, an initialization of GrowCut took about one minute. DSC: Dice Score Coefficient (%), HD: Hausdorff Distance (voxel).

| Lower Jawbone No. | Volumes of the lower Jawbones (mm$^3$) | | HD(voxel) | DSC (%) | Times (min.) |
|---|---|---|---|---|---|
| | Ground truth A | Algorithmic | | | |
| 1 | 30507.8 | 26710.9 | 29.22 | 83.26 | 36 |
| 2 | 17333 | 21200.4 | 51.39 | 80.73 | 46 |
| 3 | 19356.9 | 19033.5 | 21.35 | 82.73 | 38 |
| 4 | 46506.9 | 47028.9 | 19.65 | 88.42 | 38 |
| 5 | 39813.6 | 50087.4 | 57.46 | 80.81 | 37 |
| 6 | 30861.2 | 30118.9 | 29.1 | 87.8 | 43 |
| 7 | 45792.7 | 52090.8 | 29.45 | 86 | 38 |
| 8 | 31525.1 | 30556.2 | 49.49 | 88.27 | 36 |
| 9 | 18150.5 | 16548.2 | 19.87 | 90.33 | 38 |
| 10 | 32951.8 | 28474.4 | 28.14 | 86.28 | 36 |

https://doi.org/10.1371/journal.pone.0196378.t003

**Table 4. Direct comparison of manual slice-by-slice (Ground truth B) and Slicer-based GrowCut segmentation (algorithmic) results for ten lower jawbones via the Hausdorff Distance and the Dice Similarity Coefficient (DSC).** The last column presents the time in minutes (min.) for the manual segmentations, in contrast, an initialization of GrowCut took about one minute. DSC: Dice Score Coefficient (%), HD: Hausdorff Distance (voxel).

| Lower Jawbone No. | Volumes of the lower Jawbones (mm³) | | HD(voxel) | DSC (%) | Times (min.) |
|---|---|---|---|---|---|
| | Ground truth B | Algorithmic | | | |
| 1 | 29413.4 | 26710.9 | 27.91 | 83.6 | 40 |
| 2 | 17730.4 | 21200.4 | 50.96 | 80.66 | 40 |
| 3 | 20067.2 | 19033.5 | 20.71 | 83.77 | 39 |
| 4 | 47508.8 | 47028.9 | 19.34 | 88.69 | 38 |
| 5 | 39733 | 50087.4 | 57.46 | 80.59 | 35 |
| 6 | 31283.1 | 30118.9 | 28.86 | 88.79 | 40 |
| 7 | 45492.8 | 52090.8 | 33.65 | 86.34 | 42 |
| 8 | 32288.9 | 30556.2 | 47.84 | 87.76 | 37 |
| 9 | 18686.3 | 16548.2 | 19.34 | 89.85 | 38 |
| 10 | 31296.5 | 28474.4 | 28.25 | 87.49 | 35 |

The measurement values of the segmentations' volumes and voxels in the regression models were localized closely along the regression lines (**Figs 7 and 8**). The difference between the gradient of the regression lines in the constructed regression models were also not significant for any of the comparisons made between manual (ground truth) and semi-automatic (GrowCut) segmentation (p<0.001). The created boxplots were similar for volume and voxel values between the segmentation groups, especially between the two ground truth schemes A and B (**Figs 7 and 8**). This is also valid for the parameters DSC and HD as shown in the Tables **9** and **10**. Both parameters were similar between the algorithmic GrowCut segmentation and the manual Ground truth A and B. Again, even a smaller difference can be seen when the ground truth segmentations A and B are compared among themselves (**Figs 9 and 10**).

For a visual assessment of the performed segmentations, **Fig 11** presents the overlap of a manual ground truth (white) and a semi-automatic (gold) segmentation. Moreover, this semi-automatic segmentation has been superimposed into a 3D visualization (gold and gray). When comparing these figures a clear and more sensitive surface visualization could be achieved by the algorithmic segmentation (GrowCut) (gold), since the ground truth schemes were created manually according to the number of image slices (white). The slice-by-slice ground truth segmentation is truly visualized as a stepwise surface countering, meaning each visualized step defines one slice. Further, some semi-automatic segmentation inaccuracy occurred in a few cases in the region of the lower jaw's condyle (gold).

**Table 5. Direct comparison of the number of voxel for the manual segmentations (Ground truth A and Ground truth B) and the semi-automatic GrowCut segmentations (algorithmic) for all ten cases with the minimum (Min), maximum (Max), mean ($\mu$) values and standard deviation ($\sigma$), respectively is shown.**

| Lower Jawbone No. | Number of Voxel | | |
|---|---|---|---|
| | Ground truth A | Ground truth B | Algorithmic |
| 1 | 166749 | 160767 | 145996 |
| 2 | 118277 | 120989 | 144668 |
| 3 | 54887 | 56901 | 53970 |
| 4 | 84897 | 86726 | 85850 |
| 5 | 153211 | 152901 | 192747 |
| 6 | 96836 | 98160 | 94507 |
| 7 | 211925 | 210537 | 241072 |
| 8 | 77436 | 79312 | 75056 |
| 9 | 123856 | 127512 | 112922 |
| 10 | 103396 | 98202 | 89347 |
| Min | 54887 | 56901 | 53970 |
| Max | 211925 | 210537 | 241072 |
| $\mu \pm \sigma$ | 119147±46957.5 | 119200.7±45568.9 | 123613.5±58013.4 |

**Table 6. Summary of manual (Ground truth A, B) vs algorithmic GrowCut segmentation results, presenting: Minimum (Min), maximum (Max), mean ($\mu$) values and standard deviation ($\sigma$) for ten lower jawbones.** DSC: Dice Score Coefficient (%), HD: Hausdorff Distance (voxel).

| | Volumes of the Lower Jawbones (cm$^3$) | | | Ground truth A, B vs. Algorithmic | | | |
|---|---|---|---|---|---|---|---|
| | Ground truth A | Ground truth B | Algorithmic | DSC (%)Ground truth A \| Ground truth B | | HD (voxel)Ground truth A \| Ground truth B | |
| Min | 17.33 | 17.73 | 16.55 | 80.73 | 80.59 | 19.65 | 19.34 |
| Max | 46.51 | 47.51 | 52.09 | 90.33 | 89.85 | 57.46 | 57.46 |
| $\mu\pm\sigma$ | 31.28±10.69 | 31.35±10.59 | 32.18±13.02 | 85.46±3.38 | 85.75±3.39 | 33.51±13.98 | 33.43±13.86 |

## Discussion

The existence of computer assisted technologies based on algorithmic software segmentation is an increasing topic of interest in the medical domain 19. This is valid for many medical fields, but especially for cases of complex surgical therapeutic planning or 3D visualization of anatomical structures, in order to reduce treatment time and improve therapeutic outcome [44].

In this trial, the feasibility and practical use of an open source algorithmic software segmentation of the lower jawbone for the visualization of diagnosis and treatment planning in the cranio-maxillofacial field has been investigated (note: only the segmentation of mandible without the presence of teeth was studied). The mandible was chosen for segmentation being the biggest and strongest bone in the maxillo-facial complex consisting of compact biological hard tissue structures [45]. Furthermore does the lower jaw represent with about 40% the highest occurrence of all facial fractures in the cranio-maxillofacial field and is moreover often involved in trauma injury due to traffic and sport accidents or violent crimes [46] that lead to time consuming preoperative planning procedures or complex surgical treatment strategies [11].

For the investigation of the open source algorithm, selected data sets of lower jawbones have been segmented manually on a slice-by-slice basis (ground truth) by clinical experts and semi-automatically with a cellular automata by the GrowCut algorithm in a retrospective, randomized, controlled trial. The segmentations have been compared successively to assess the degree of agreement and accuracy by defined parameters (segmentation time, volume, voxel, DSC, HD). Additionally, pure generated ground truth segmentations have been compared amongst each other to determine any uncertainty in the control group data sets.

**Table 7. Difference values (pared t-test, p) and product-moment correlation (Pearson, r) for volume comparisons of the manual (Ground truth A, B) and interactive GrowCut (algorithmic) segmentations are shown.** No statistical significance (p<0.05) was observed between the segmented volumes and a high direct proportional correlation (r) close to the values one can be seen between the segmented volumes.

| Volume comparison | Significance (p) | Coefficient (r) |
|---|---|---|
| Ground truth A to Ground truth B | p = 0.803 | r = 0.997 |
| Ground truth A to Algorithmic | p = 0.550 | r = 0.943 |
| Ground truth B to Algorithmic | p = 0.571 | r = 0.948 |

**Table 8. Difference values (paired t-test, p) and product-moment correlation (Pearson, r) for the comparisons of voxel units of the manual (Ground truth A, B) and interactive GrowCut (algorithmic) segmentations are shown.** No statistical significance (p<0.05) was observed between the segmented volumes and a high direct proportional correlation (r) close to the value one can be seen between the segmented volumes.

| Voxel comparison | Significance (p) | Coefficient (r) |
|---|---|---|
| Ground truth A to Ground truth B | p = 0.960 | r = 0.998 |
| Ground truth A to Algorithmic | p = 0.502 | r = 0.948 |
| Ground truth B to Algorithmic | p = 0.493 | r = 0.957 |

DSC and HD values are known to be valid parameters for assessing the overlap and agreement of two segmented volumes [38–40] and were already used by others to assess interactive segmentation processes of e.g. glioblastoma multiforma in the brain [47].

Regarding to the literature there are some articles that evaluate the accuracy of algorithmic software segmentation of 3D models or of computer-aided 3D surface reconstructions from CT-based DICOM image files [48–54]. However, many of these authors assess the accuracy of the varying segmentation approaches by comparing virtual segmentation volumes with measurements performed on real manufactured objects of the same structure. More precisely, these authors compare the virtuality of structures to their reality or the structure's reality to their virtuality such as in comparisons between virtual segmentations and real 3D printed models or manufactured phantoms and virtual 3D scans. Moreover just a few articles really concerned the cranio-maxillofacial region, when those only focusing the dental medical field are excluded [33, 54].

However, none of the existing articles dealing with the assessment of segmentation accuracy in the cranio-maxillofacial field compared the used segmentation approach based on lower jaw CT-images to the ground truth volumes of the same anatomical structure, which directly shows any occurring variability in the segmentation process and provides therefore high accurate results in the assessment procedure.

The difficulty and high effort in the creation of ground truth schemes may be a reason for the low existence of these volumes as controlled data samples for the comparison of segmentation approaches with special regard to already existing algorithms.

In this trail, the randomly selected data-sets were chosen because of the 1) frequent involvement of the lower jaw in complex trauma cases and in parallel because of 2) the solid anatomy of the bone. This both leads 1) to the need of frequent clinically relevant surgical interventions in the clinical practice such as complex osteosynthesis or lower jaw reconstructions 46 and provides 2) the opportunity of an objective comparison between a segmentation method and the created ground truth volumes. As performed in this trial manual slice-by-slice segmentation–for the generation of ground truth volumes as control samples—was also done by Szymor et al. and Yan-Hui Sang et al. to assess the accuracy of software segmentation or 3D surface reconstructions in their study [33, 48]. Szymor et al. evaluated the segmentation accuracy of parts of the inner orbital wall by comparing the segmentation approach with the 3D printed models of the same structure. They stated that the use of 3D Slicer–that was also used as a software platform in this trial–is accurate enough to create models for a clinical use [33]. However such inter-dimensional study designs could lead to comparison inaccuracies influenced by many involving factors such as the used printer, the printer's own software, the varying materials for the 3D model production or the variability in the measurements between the manufactured 3D models and the virtual software-based segmentation approach.

Others working in the field of (semi-) automatic jawbone segmentation are Barandiaran et al. [55] who presents the automatic segmentation and reconstruction of mandibular structures from Computed Tomography (CT) data. For the automatic segmentation process they establish a pipeline consisting of several threshold filters. Amongst others, they apply the multiple threshold method by Otsu [56]. Harandi et al. [57] introduce upper and lower jaw segmentation in dental x-ray images using a modified Active Contour [58–61]. In a first step, they separate the upper and lower jaw, followed by a modified geodesic active contour and morphological operations. The automatic segmentation of mandibles in low-dose CT data is demonstrated by Lamecker et al. [62]. For an automatic segmentation in low-dose images, their work explores the ability of a model-based segmentation using a 3D statistical mandible model. The method consists of a training and a segmentation phase and includes a deformation strategy for detecting the mandibular bone. A segmentation approach to extract the trabecular jawbone in cone beam CT (CBCT) data sets is studied by Nackaerts et al. [63]. In summary, they used adaptive thresholding for the automatic segmentation of upper and lower jaws. For testing two volumes of interest each jaw were manually delineated and micro-CT images served as high-resolution ground truth images. Tan et al. [64] present threshold segmentations in 3D reconstructions of mandible CT images. To obtain an approximate segmentation result they used dilation operations, and a more precise segmentation results was achieved with the additional help of logical operations and region growing. Kainmueller and colleagues 23 performed the automatic extraction of mandibular nerve and bone from cone-beam CT data. The fully-automatic method is based on a combined statistical shape model [65, 66] of the nerve and the bone and a Dijkstra-based [67, 68] optimization procedure. Furthermore, Koningsveld [69] presents the automated segmentation of the mandibular nerve canal in CBCT images in his thesis. The approach begins with a combination of a smoothing and gradient filter to reduce noise and enhance the edges of the canal, which prepares the image for a fuzzy-connectedness method. Finally, the results are interpolated to fill in gaps and correct any errors.

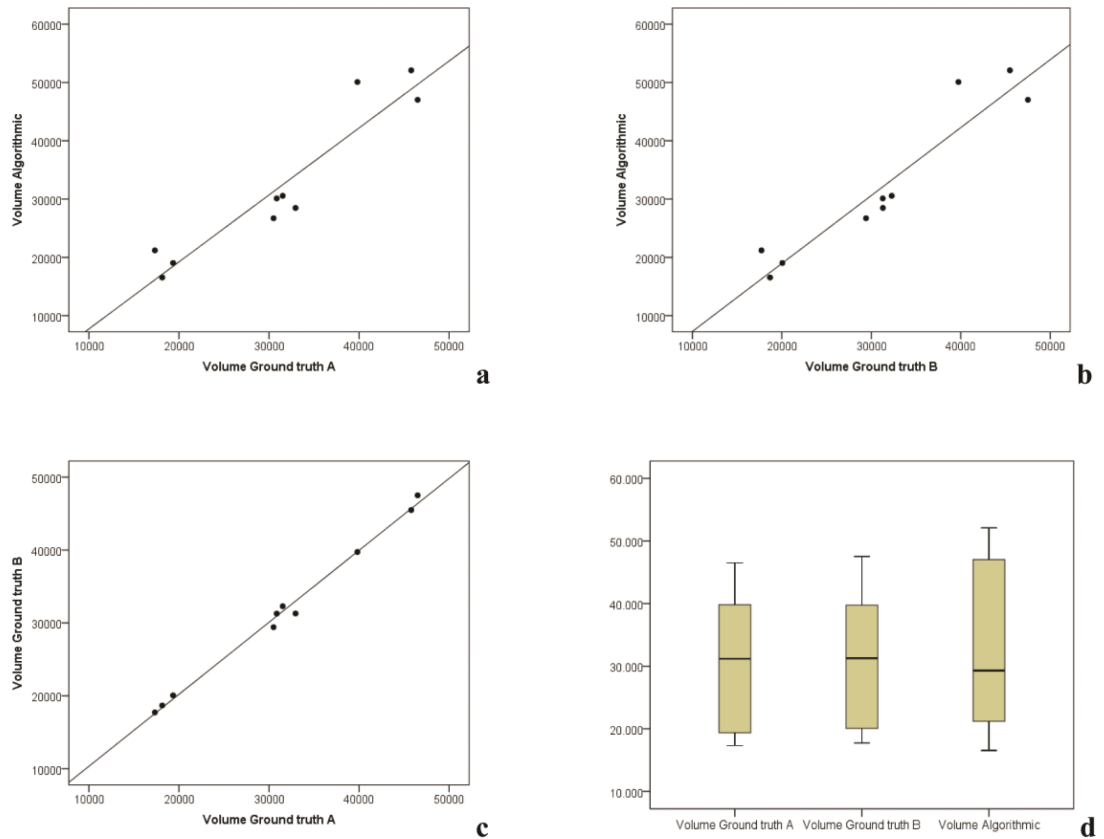

**Fig 7.** Regression analysis of volume measurements: Volume measurement distributions are similar along the regression lines between the manual (Ground truth A, B) and GrowCut segmentations (**a, b**). Moreover volume measurement distributions are closely located along their regression lines when comparing manual and semi-automatic segmentation volumes (**a, b, c**). Volume measurements of the segmentations are further shown in a boxplot diagram (**d**), providing nearly equally ground truth segmentation volumes. Difference values between the gradient of the regression lines were not statistically significant (p<0.001). *Note: volumes are given mm$^3$ in the tables.*

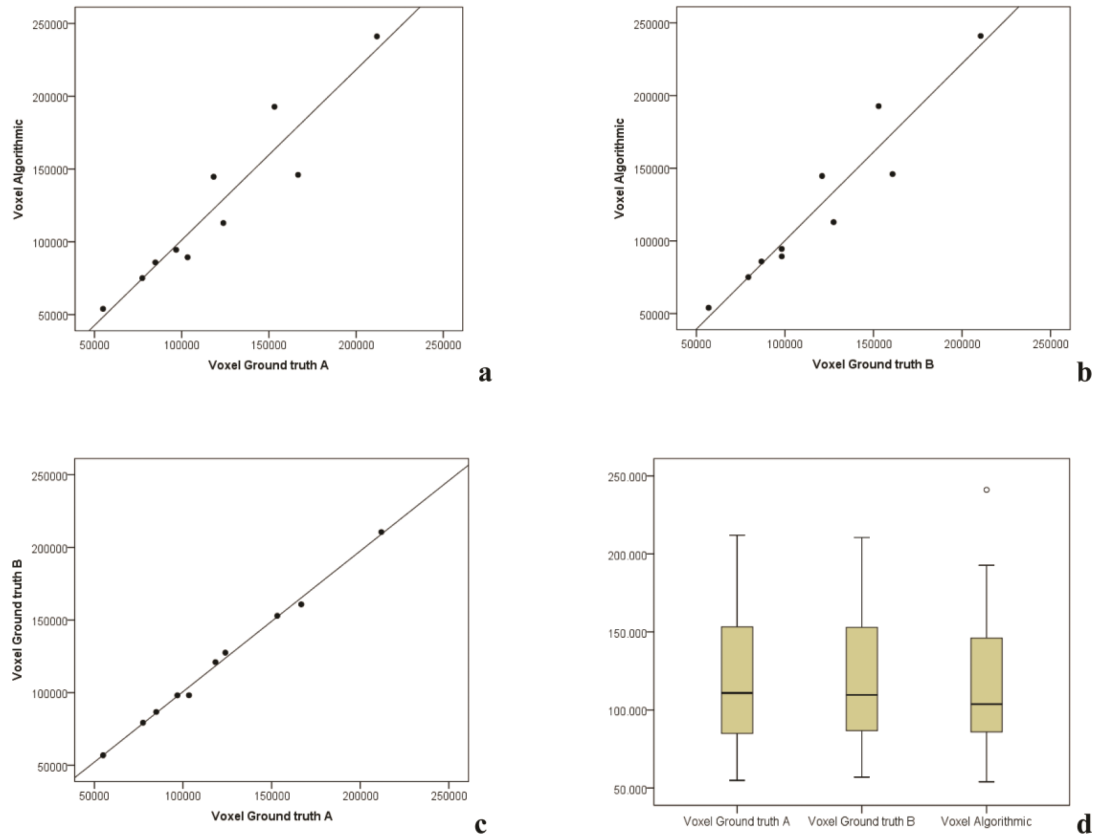

**Fig 8.** Regression analysis of voxel measurements: Voxel measurement distributions are similar along the regression lines between the manual (Ground truth A, B) and GrowCut segmentations (**a, b**). Moreover voxel measurement distributions are closely located along their regression lines when comparing manual and semi-automatic segmentation volumes (**a, b, c**). Voxel measurements of the segmentations are further shown in a boxplot diagram (**d**), providing nearly equally ground truth segmentation volumes. Difference values between the gradient of the regression lines were not statistically significant (p<0.001).

In this trial the user marked parts of the mandibular bone and the surrounding background in an axial, a sagittal and a coronal slice in order to perform the semi-automatic GrowCut segmentation. This course of action is related to cellular automata based segmentations of brain tumors in earlier studies [70]. Equivalent to the previous studies, a user could accomplish an initialization of the algorithm in approximately a minute. After the initialization, the automatic segmentation of the whole lower jawbone can already been started without any further user interactions or parameter settings.

**Table 9. Comparisons of the DiceScore Coefficients (DSC, %) including minimum (Min), maximum (Max), mean ($\mu$) values and standard deviations ($\sigma$) are shown.**

| DSC (%) comparison | Min | Max | $\mu$ | $\sigma$ |
|---|---|---|---|---|
| Ground truth A to Ground truth B | 91.7 | 94.7 | 93.6 | 1.0 |
| Ground truth A to Algorithmic | 80.7 | 90.3 | 85.5 | 3.4 |
| Ground truth B to Algorithmic | 80.6 | 89.9 | 85.8 | 3.4 |

**Table 10. Comparisons of Hausdorff Distances (HD, voxel) including minimum (Min), maximum (Max), mean (μ) values and standard deviations (σ) are shown.**

| HD (voxel) comparison | Min | Max | μ | σ |
|---|---|---|---|---|
| Ground truth A to Ground truth B | 2.2 | 6.3 | 4.0 | 1.2 |
| Ground truth A to Algorithmic | 19.7 | 57.5 | 33.5 | 14.0 |
| Ground truth B to Algorithmic | 19.3 | 57.5 | 33.4 | 13.9 |

According to our results, functional stable and statistical satisfying qualitative and quantitative segmentation outcomes could be produced with interactive algorithmic support in a short time of approximately one minute. More precise, average Dice Score coefficient values of over 85%, Hausdorff Distances below 33.5 voxel while in parallel similar voxel and volume values were achieved, when comparing the semi-automatic with the manual ground truth segmentations, which proofs that Grow-Cut is able to come up with high segmentation accuracy according to the segmented anatomical structures. Hence, the Dice Score coefficient values and Hausdorff Distances of this study may be acceptable for a clinical relevant use for fast and easy 1) 3D visualization, 2) 3D model printing or 3) template creation for the preoperative orientation of ostheosynthesis material adaption. According to the clinical experience Dice Score coefficient values of over 80% are in general sufficient for the mentioned clinical relevant use. Moreover these parameters in our study were similar to the overlap values observed in semi-automatic segmentation processes with GrowCut in assessing volumetries of glioblastoma mulitforma [47]. Additionally, volume and voxel values of the semi-automatic segmentation in this trial were not significantly different compared to the ground truth segmentations. Also, Pearsons's correlation coefficient calculated for segmentation volumes and the number of voxel was close to the value one, which shows a high direct, positive correlation and accordance between the GrowCut segmentation group and the ground truth control group. When comparing the more time consuming two ground segmentations–over 38 minutes on average–generated by two clinical specialists among each other, Dice Score Coefficient values were even higher and Hausdorff Distances even smaller than those between GrowCut segmentations and ground truth volumes. Again, both, neither volume values nor voxel units of the manual segmentations were significantly different when the two ground truth schemes were compared. In order high positive, direct correlations could be observed showing a high similarity between the compared ground truth volumes.

Further, the created regression analysis support these findings showing that the constructed regression lines between the GrowCut segmentation and ground truth control group according to volumes and voxel were not significantly different for any comparison made between the performed segmentations (p<0.01).

When considering these results, this trial shows an efficient course of action in lower jawbone volumetry by an effective, easy available approach of semi-automatic segmentation using GrowCut when directly compared to the manually ground truth segmentation. Further, this trial shows that the used ground truth schemes were nearly identical by achieving a very high degree of overlap, although they were generated independently by two clinical experts. Albeit the ground-truth schemes originate from a strictly defined data-set selection, these volumes can be used as an objective, valid control group data sample without bias causing variability or comparison inaccuracy due to manual slice-by-slice segmentation.

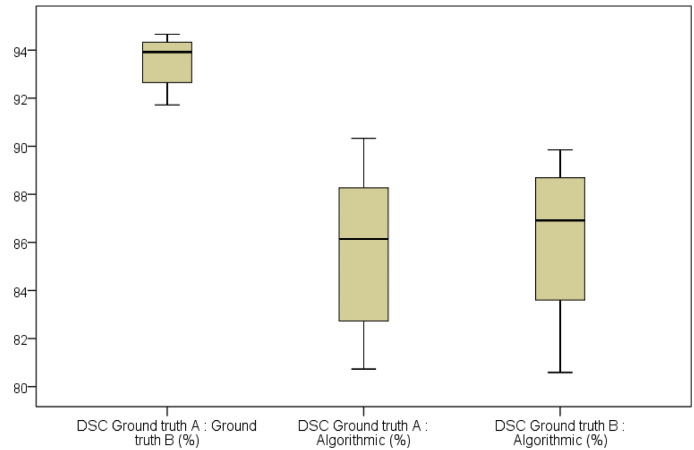

**Fig 9. Comparisons of the DiceScore Coefficients (DSC, %) are shown in a boxplot diagram.** Plots between manual (Ground truth A, B) and GrowCut (Algorithmic) segmentations are similar, providing valid ground truth control samples created independently by two clinical experts (A, B).

Referring to the presented results, the hypothesis of this trail can be confirmed that the accuracy and accordance of a commonly available and licensed free semi-automatic open-source segmentation algorithm is not significantly different than the ground truth data of the same anatomy. The hypothesis' confirmation is based on the following findings: 1) missing significant differences between the compared segmentations, 2) high degrees of overlap between all segmented volumes, 3) functional stable segmentation processes, 4) valid ground truth control samples and 5) low time consumption needed for the investigated semi-automatic segmentation approach. Therefore, these findings can suggest the algorithm's clinical practicability.

Despite these results we are aware of some limitations concerning this trial: First, the segmentation method assessed in this article is not completely new, since the already existing GrowCut algorithm can be used in many variations in different software applications and platforms [70]. Second, lower jaw segmentation has also been previously carried out by other groups including image-based mandibular nerve extraction [23] and is described within a pilot project in combination with a computer-aided trauma simulation system by haptic feedback [11]. Third, although the data sets used in this trial were selected randomized in the clinical routine, a higher amount of data-samples would probably have more impact in assessing feasibility and accuracy of the used segmentation approach. Fourth, although ground truth generation was tried to be performed as valid as possible by two clinical experts and were proofed as valid control samples by analytical statistical calculations, a real image-based ground truth scheme used as comparative segmentation volume is impossible to create, since every segmentation approach has to relief on certain image-based landmarks. Fifth, some inaccuracy was observed when segmenting the lower jaw's condyle, since this region is physiologically overlapped by the skull base and strongly interferes with other anatomical structures. Sixth, we did not include the segmentation of teeth within our medical data-sets. Since these data-sets often lead to image-based artifacts and generate incomplete or inaccurate CT-scans, they were excluded in the data selection process to obtain a qualitative comparative assessment. In more detail, incomplete data sets can frequently include radiological artifacts e.g. from metal tooth parts or ostheosynthesis materials, include free pieces of bone structures with missing bone contour anatomy as it is the case when pathological data sets are used, or are inaccurate due to missing or damaged slices. Since the hypothesis of this study was defined to compare an open-source segmentation algorithm with the generated ground truth data of the same anatomy, only physiologic data sets with clear bone contours were used. Incomplete data sets such as pathological data sets or data sets with missing or damaged slices would have affected the ground truth generation, which was done by medical specialists due to strong occurring subjectivity in the manual segmentation process. For an adequate objectivity in the assessment of the investigated open-source algorithm, we needed an accurate and clearly generated valid control group of the same anatomical structures (ground truth). Therefore, this could be only done when physiologic data sets with clear bone contours and anatomical structures without artifacts are used.

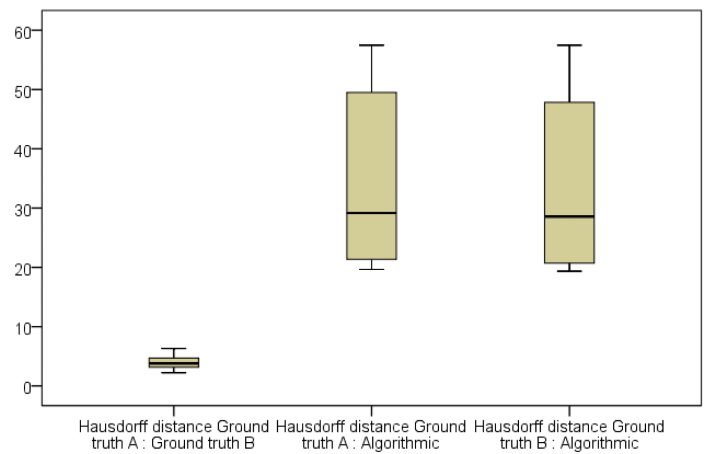

**Fig 10. Comparisons of Hausdorff Distances (HD, voxel) are shown in a boxplot diagram.** Plots between manual (Ground truth A, B) and GrowCut (Algorithmic) segmentations are similar, providing valid ground truth control samples created independently by two clinical experts (A, B).

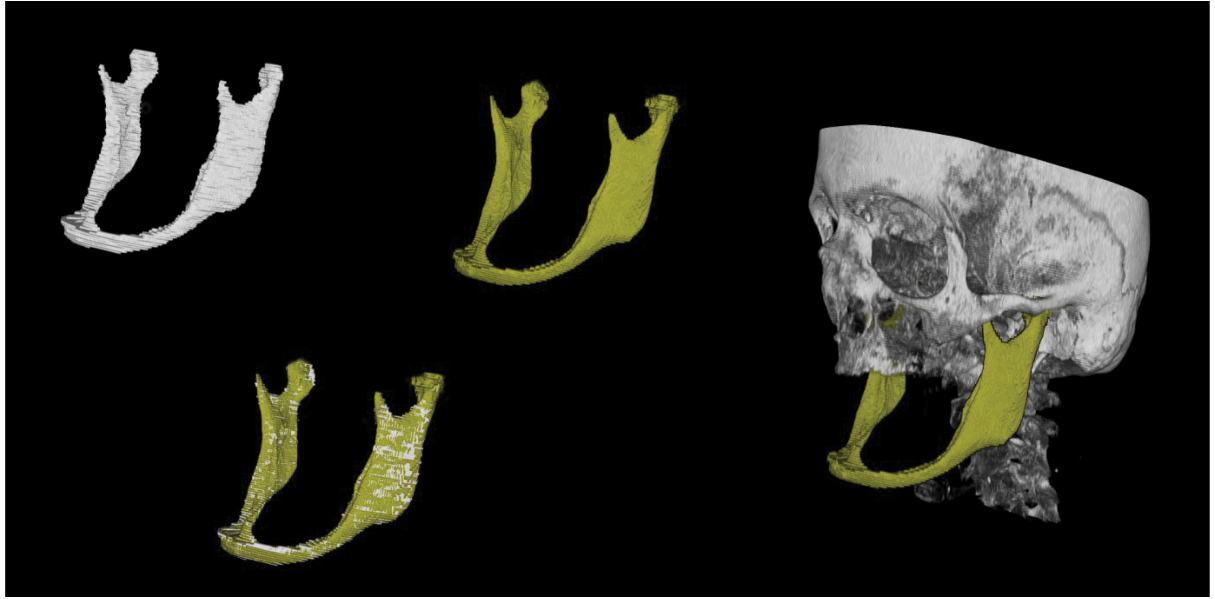

**Fig 11. Lower jaw semi-automatic segmentation: A ground truth scheme (white) and an semi-automatic segmentation (GrowCut, gold) is shown, including an overlay for accuracy and accordance assessment of the open-source algorithm (GrowCut) and a superimposed visualization of the semi-automatic segmentation (gold) into a 3D visualization of the patient's skull (gray).** The upper jaw is faded out for a clear visualization. *Note: The segmented mandible is–due to missing teeth–strongly atrophied. The thin bone is well visualized by the semi-automatic algorithm (GrowCut). Some semi-automatic inaccuracy can be observed at the mandible's condyle.*

Additionally, to the best of the authors' knowledge this study is the first that compares an open-source algorithm using data of the lower jaw with the ground truth data of the clinical routine. Therefore, this study is more or less a pilot project. In that context the study was performed to gain first results in this field and generate an objective control group—which was done by the ground truth generation of the used data sets—that can be used for the assessment of the investigated algorithm. However, there are several areas for a future work, like the evaluation of the investigated GrowCut algorithm with a greater amount of medical data. As a soon future work project we will take these medical data—such as incomplete data sets—into account and further also include a systematic comparison with other freely available segmentation algorithms.

Moreover, we want to state that no CE-marked software (e.g. Brainlab, Materialise CMF Module, Maxilim, IPS Case Designer) was validated in the study although CE-marked software is becoming increasingly important. However with this investigation we did not want to set a new gold standard by evaluating an open-source algorithm. More we wanted to point out a different solution with a free and easily available medical image-based analysis, since the Slicer based GrowCut algorithm was not objectively investigated yet by using ground truth data generated by medical specialists in the lower jaw. In any case, CE-marked software is the gold standard in head and neck surgery departments, but only if these software packages are available at the clinical center. Still numerous departments do not work with these CE-marked software packages since the packages are expensive and usually need additional human resources for their use. Also at our department CE-marked software is difficult to handle since the updates and the use are directly connected to strong monetary aspects. In the clinical practice, especially in smaller head and neck departments, software without CE-marks for the segmentation processes for 3D reconstructions, visualizations or the preparation of 3D printable models is routinely used to avoid additional financial costs or working time aspects. These software packages are usually functional stable, easily and freely available and offer many options for medical image-based analysis.

In our study we selected the 3D Slicer because this software is license-free and easy available, and offers multiple tools for image processing. Further, the software can be used in every clinical center independent from financial aspects. This is especially relevant for smaller clinical departments where monetary aspects are an issue concerning image-processing, 3D visualization or 3D printing. Limitations of the software can be seen in missing hardware or computer power that is needed. Additionally, the software does not have a CE mark or ISO certification, Hence it would not be allowed to print a Slicer segmentation based ostheosynthesis material for an intraoperative use. However the software may be clinically used for virtual 3D visualization, 3D model printing for macroscopic operation or resection planning and/or for the creation of templates for the preoperative orientation of ostheosynthesis material adaption. Taking these points and the results of our study into account the investigated method and image processing tools may provide accurate segmentation results of adequate quality when clinically used as mentioned above, although the software does not have a medical ISO certification.

Further the results presented in this article are in accordance to other works that focus on the assessment of varying segmentation approaches based on 3D model production [33], 49–53 and further with the results from other authors that used the same algorithm in other medical fields such as for the segmentation of tumors or cDNA [71–73].

In summary, complete functional stable and time saving segmentation could be performed by the interactive user guided approach. Additionally, advantages of the used method are 1) a free access to the software and the prototyping platform, 2) the avoidance of licensed based monetary services e.g. outsourced services or acquisition of monetary software packages, 3) the opportunity of controlled testing series by other centers and 4) a further development by other groups or specialists. Thus, supporting tasks like the planning of operations in maxillofacial surgery, the visualization of treatment strategies in complex surgical cases or the production of 3D models of the mandible could possibly be performed within a clinical center, which would lead to a shortened operative planning and treatment time while in parallel improve the treatment quality [19, 55].

In order, the achieved research highlights of the presented work are:

- Manual slice-by-slice segmentations of mandibles have been performed by clinical experts to obtain ground truth of the lower jawbone boundaries and estimates of rater variability.

- Users have been trained in segmenting mandibles with GrowCut and the Editor module of 3D Slicer.

- Trained users used Slicer to segment a mandibular evaluation set.

- Algorithmic segmentation times have been measured for GrowCut based segmentation in 3D Slicer.

- Quality evaluation of the segmentations and comparison has been performed by calculating the Dice Similarity Coefficients (DSC), Hausdorff Distances (HD) and included statistical analysis to directly compare volumes, voxel, DSC and HD values of the segmentation approach, and the ground truth of the same anatomy.

- The feasibility and practical assessment of an open source segmentation software was performed in an evaluation that has not been described before at a center of cranio-maxillofacial surgery to be further possibly used in the clinical practice.

- A unique image data set collection of the lower jaw bone and two manual ground truth segmentations are provided for a further assessment or development of varying segmentation approaches or own research purposes (please see the acknowledgments section for more information).

As mentioned above there are several areas for future work, like the evaluation of the Grow-Cut algorithm with a greater amount of medical data and a systematic comparison with other freely available segmentation algorithms, like the robust statistics segmentation (RSS). Further, analyze the stability of the tool under different background noise, like the presence of artifacts, for instance, the metallic artifacts. In addition, we plan to expand the study to perform the segmentation also on a CE-marked software package and the segmentation and validation of other facial anatomical regions. Additionally, the segmentation results, achieved within this work, can support the computer-aided reconstruction of facial defects and surgical template design for oral implantology [21]. Furthermore, the results can be imported into a medical augmented reality (AR) system for surgical navigation (www.augmentedrealitybook.org) [74], an Virtual Reality (VR) environment [75, 76] and an optical see-through head-mounted display (HMD) [77] to be used e.g. for the resection of tumors or complex surgical cases in cranio-maxillofacial and head and neck surgery. Finally, we want to use the segmentations for the computer-aided reconstruction of facial defects with miniplates [78].

## Additional information

A few initial results of this trial have been presented and discussed as a talk [79] at the 20<sup>th</sup> *Annual Congress of the Austrian Society of Oral and Maxillofacial Surgery (ÖGMKG), in Bad*

*Hofgastein, Salzburg, Austria* and as a late breaking research poster [80] at the *38<sup>th</sup> Annual International Conference of the IEEE Engineering in Medicine and Biology Society (EMBC)* in *Orlando, FL, USA.* However, at the *ÖGMKG congress* we showed only some first outcomes of the segmentation results and at the *EMBC* we presented only a one page summarized description of the algorithm. All statistical results and a precise description of the methods are only presented in full details within this contribution.

## Acknowledgments

This work received funding from the Austrian Science Fund (FWF) KLI 678-B31: "enFaced: Virtual and Augmented Reality Training and Navigation Module for 3D-Printed Facial Defect Reconstructions" (PIs: Jürgen Wallner and Jan Egger), BioTechMed-Graz in Austria ("Hardware accelerated intelligent medical imaging"), the 6<sup>th</sup> Call of the Initial Funding Program from the Research & Technology House (F&T-Haus) at the Graz University of Technology (PI: Jan Egger) and the TU Graz Lead Project ("Mechanics, Modeling and Simulation of Aortic Dissection"). Dr. Xiaojun Chen receives support from the National Key Research and Development Program of China (2017YFB1104100, 2017YFB1302903), Foundation of Science and Technology Commission of Shanghai Municipality (15510722200, 16441908400), and Shanghai Jiao Tong University Foundation on Medical and Technological Joint Science Research (YG2016ZD01, YG2015MS26). Further, our unique image data set collection of the lower jaw bone and the two manual ground truth segmentations can be requested from the authors. Finally, a step-by-step video demonstrating the segmentation process can be found under the following YouTube channel: https://www.youtube.com/c/JanEgger.

## Author Contributions

**Conceptualization:** Jürgen Wallner, Jan Egger.

**Data curation:** Jürgen Wallner.

**Formal analysis:** Jürgen Wallner, Kerstin Hochegger, Irene Mischak, Jan Egger.

**Funding acquisition:** Dieter Schmalstieg, Jan Egger.

**Investigation:** Kerstin Hochegger, Jan Egger.

**Methodology:** Jürgen Wallner, Jan Egger.

**Project administration:** Dieter Schmalstieg, Jan Egger.

**Resources:** Jürgen Wallner, Xiaojun Chen, Knut Reinbacher, Mauro Pau, Tomislav Zrnc, Katja Schwenzer-Zimmerer, Wolfgang Zemann, Dieter Schmalstieg, Jan Egger.

**Software:** Kerstin Hochegger, Jan Egger.

**Supervision:** Katja Schwenzer-Zimmerer, Dieter Schmalstieg, Jan Egger.

**Validation:** Jürgen Wallner, Jan Egger.

**Visualization:** Jürgen Wallner, Jan Egger.

**Writing – original draft:** Jürgen Wallner, Jan Egger.

**Writing – review & editing:** Jürgen Wallner, Jan Egger.